%% file: main.tex
\documentclass[10pt,twocolumn,letterpaper]{article}

\usepackage[pagenumbers]{cvpr} 

\input{preamble}

\definecolor{cvprblue}{rgb}{0.21,0.49,0.74}
\usepackage[pagebackref,breaklinks,colorlinks,citecolor=cvprblue]{hyperref}

\title{MS-CLR: Multi-Skeleton Contrastive Learning for Human Action Recognition}

\author{
  Mert Kiray$^{1,2}$\thanks{Authors contributed equally to this work.} \quad
  Alvaro Ritter$^{1}$\footnotemark[1] \quad
  Nassir Navab$^{1}$ \quad
  Benjamin Busam$^{1,2}$ \\[1ex]
  $^{1}$Technical University of Munich \\
  $^{2}$3Dwe.ai
}

\begin{document}
\input{figures/teaser/teaser}
\maketitle
\input{sec/0_abstract}    
\input{sec/1_intro}
\input{sec/2_related}
\input{sec/3_method}
\input{sec/4_experiments}

\input{sec/5_conclusion}
{
    \small
    \bibliographystyle{ieeenat_fullname}
    \bibliography{main}
}

\end{document}

%% file: preamble.tex
\usepackage[dvipsnames]{xcolor}

\newcommand{\customddag}{\textcolor{black}{\ddag}}

\usepackage{lipsum}
\usepackage{tabularx}
\usepackage{float}   

%% file: figures/teaser/teaser.tex
\makeatletter
\g@addto@macro\@maketitle{
  \vspace{-30pt}
  \begin{figure}[H]
  \setlength{\linewidth}{\textwidth}
  \setlength{\hsize}{\textwidth}
  \centering
    \includegraphics[width=\linewidth]{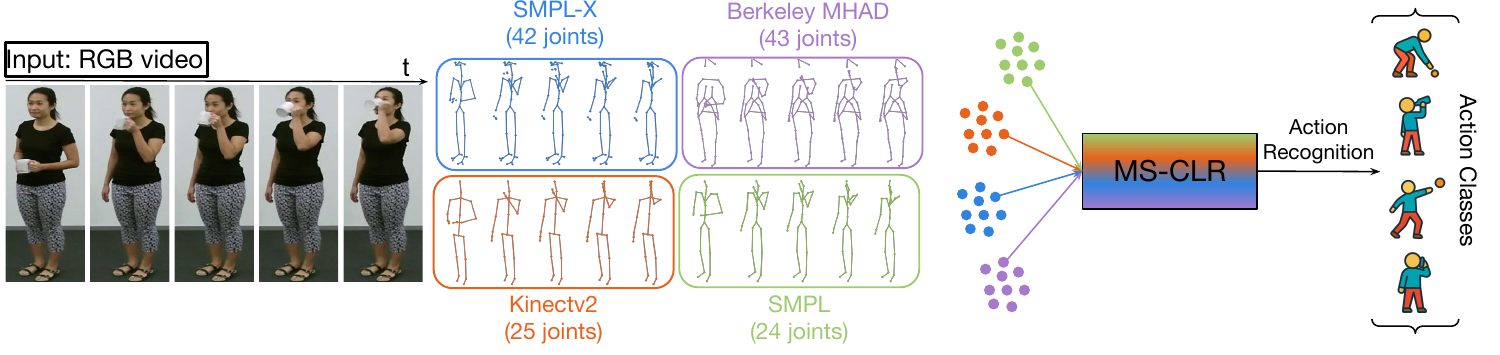}
    \vspace{-20pt}
    \caption{MS-CLR leverages multiple skeleton conventions extracted from the same video to provide structurally diverse views for contrastive learning. By integrating anatomically distinct pose formats within a unified graph-based framework, MS-CLR encourages the model to learn structure-aware invariances, improving robustness and generalization in action recognition.}
    \label{fig:teaser}
	\end{figure}
}
\makeatother

%% file: sec/0_abstract.tex
\begin{abstract}
\vspace{-5pt}
Contrastive learning has gained significant attention in skeleton-based action recognition for its ability to learn robust representations from unlabeled data. However, existing methods rely on a single skeleton convention, which limits their ability to generalize across datasets with diverse joint structures and anatomical coverage. We propose Multi-Skeleton Contrastive Learning (MS-CLR), a general self-supervised framework that aligns pose representations across multiple skeleton conventions extracted from the same sequence. This encourages the model to learn structural invariances and capture diverse anatomical cues, resulting in more expressive and generalizable features. To support this, we adapt the ST-GCN architecture to handle skeletons with varying joint layouts and scales through a unified representation scheme. Experiments on the NTU RGB+D 60 and 120 datasets demonstrate that MS-CLR consistently improves performance over strong single-skeleton contrastive learning baselines. A multi-skeleton ensemble further boosts performance, setting new state-of-the-art results on both datasets. Code available at \href{https://3dwe-ai.github.io/ms-clr}{3dwe-ai.github.io/ms-clr}.
\end{abstract}

%% file: sec/1_intro.tex
\vspace{-10pt}
\section{Introduction}
Human action recognition is a longstanding task in computer vision with applications in human-computer interaction~\cite{zhang2012microsoft}, autonomous driving~\cite{grigorescu2020survey, martin2019drive}, sports analytics~\cite{bialkowski2014large, soomro2015action}, and surveillance~\cite{mabrouk2018abnormal}. Despite substantial progress, the task remains challenging due to the complex spatial and temporal dependencies and subtle motion variations in human behavior.

In recent years, skeleton-based methods have gained popularity due to their compactness, interpretability, and robustness to appearance variation. However, many existing models rely heavily on large, labeled datasets, which limits their scalability and generalization. Furthermore, privacy concerns associated with RGB data have led to growing interest in skeleton-only action recognition~\cite{pareek2021survey}.

Self-supervised contrastive learning offers a promising alternative by leveraging unlabeled data to learn transferable motion representations~\cite{chen2020simple, qian2021spatiotemporal}. While it has been effective in image and video domains, adapting contrastive learning to skeleton-based action recognition introduces new challenges. Most existing methods use a single skeleton format and depend on handcrafted augmentations~\cite{thoker2021skeleton, guo2022contrastive, zhang2023hierarchical}, which limits their ability to generalize across datasets with different joint layouts, resolutions, and anatomical coverage.

We address this gap by proposing Multi-Skeleton Contrastive Learning (MS-CLR), a self-supervised learning strategy that aligns pose representations across multiple skeleton conventions extracted from the same video. Unlike traditional augmentation-based contrastive methods, MS-CLR leverages the structural diversity present in different pose representations. For example, formats such as SMPL-X~\cite{pavlakos2019expressive}, Kinectv2~\cite{zhang2012microsoft}, and Berkeley MHAD~\cite{ofli2013berkeley} differ significantly in joint count and anatomical coverage, and this diversity serves as a supervisory signal that improves robustness to variation in body topology. While these formats are used in our experiments, MS-CLR is compatible with any skeleton convention that can be represented as a spatiotemporal graph.

To support this approach, we adapt the Spatiotemporal Graph Convolutional Network (ST-GCN)~\cite{yan2018spatial} to process multiple skeleton formats by mapping all skeleton formats into a shared input space while preserving their native graph structure. MS-CLR operates within this unified structure during pretraining and requires no architecture change at test time. As illustrated in Fig.~\ref{fig:teaser}, our framework uses multiple structurally distinct skeletons extracted from the same video to provide diverse views for contrastive training. Additionally, we show that multi-skeleton classifier ensembling provides further gains by combining complementary pose-specific representations.

\textbf{Contributions.} The main contributions of this paper are:
\begin{itemize}
    \item We propose \textbf{MS-CLR}, a self-supervised contrastive learning framework that exploits structural diversity in human pose representations by aligning multiple skeleton conventions of the same sequence.
    \item We introduce a \textbf{multi-skeleton ST-GCN} backbone that unifies diverse joint layouts using format-specific graphs and joint-aligned input representations.
    \item We demonstrate consistent gains over strong single-skeleton contrastive baselines on NTU RGB+D 60 and 120, with additional performance boosts from multi-skeleton classifier ensembling.
\end{itemize}

%% file: sec/2_related.tex
\begin{figure*}[ht!]
    \centering
    \includegraphics[width=\textwidth]{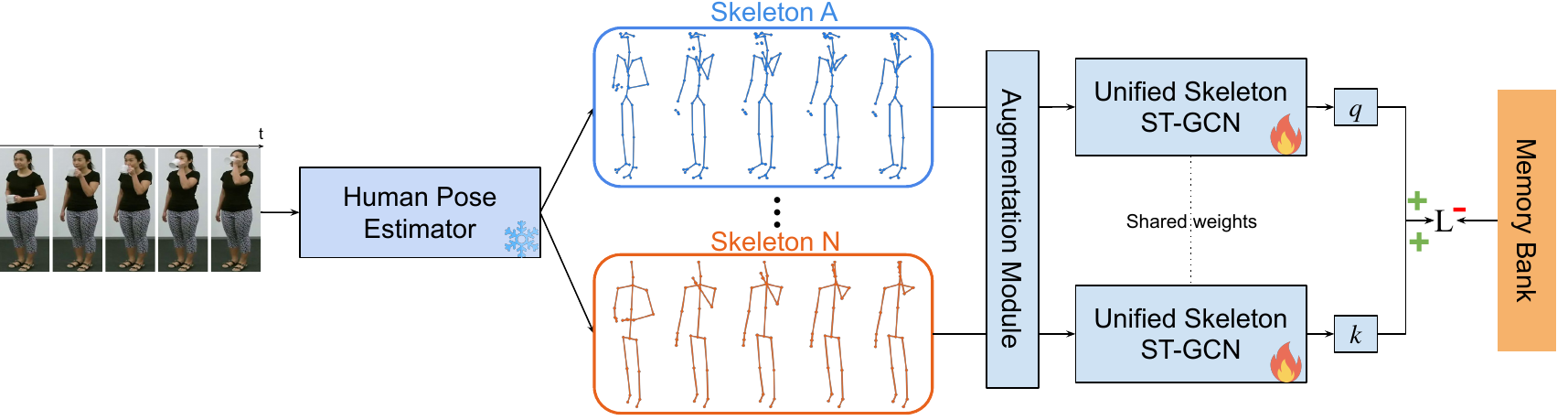}
    \caption{Overview of MS-CLR: The figure illustrates the integration of multiple skeleton conventions into the contrastive learning pipeline. Different skeleton formats are extracted using a pretrained human pose extractor that is capable of extracting multiple different skeleton formats from an RGB input. The resulting skeletons are processed through a unified ST-GCN, and features are contrasted using a multi-skeleton contrastive loss to learn structure-aware representations for action recognition.}
    \label{fig:method_overview}
\end{figure*}

\section{Related Work}

\subsection{3D Pose Estimation from Images}
3D pose estimation determines the three-dimensional position and orientation of human body parts from visual data, typically captured by RGB cameras. These methods predict keypoints on major body joints to reconstruct human skeleton motion. In modern 3D avatars~\cite{jung2023deformable,zielonka2023drivable,qian20243dgs}, the skeleton drives the movement of volumetric models, enabling realistic motion. Understanding how the skeleton controls movement is fundamental to models like OpenPose~\cite{Cao_2017_CVPR} and VNect~\cite{10.1145/3072959.3073596}, which map 2D keypoints into 3D structures. However, these methods are limited by their reliance on specific skeleton formats, reducing their scalability across datasets.

MeTRAbs~\cite{Sarandi_2023_WACV} addressed this limitation by introducing an autoencoder-based method that bridges different skeleton formats, ensuring consistent 3D pose representations. Building on this idea, our method applies multi-format consistency to the domain of action recognition by handling multiple skeleton formats, thereby improving scalability and accuracy across diverse datasets. This extension not only maintains consistency but also enhances the model's ability to capture the complex spatio-temporal dynamics essential for accurate human action recognition.

\subsection{Skeleton-Based Action Recognition from Video}
Recent advancements in skeleton-based action recognition leverage CNNs, LSTMs, and GCNs to handle the spatial and temporal aspects of skeleton data. Early methods based on hand-crafted features~\cite{wang2012mining, Vemulapalli_2014_CVPR, Vemulapalli_2016_CVPR} have evolved with the introduction of deep learning techniques. Methods using RNNs~\cite{Du_2015_CVPR, song2018spatio, zhang2019view} and CNNs~\cite{du2015skeleton, Ke_2017_CVPR, liu2017enhanced} have shown significant improvements. CNNs, applied to skeletal data formatted as pseudo-images, excel in extracting spatial features. Techniques like JointCNN~\cite{wang2022joint} and 3D-CNNs like PoseConv3D~\cite{Duan_2022_CVPR} effectively address temporal dynamics. LSTMs manage temporal aspects, improving the handling of long-term dependencies and noise through mechanisms like context-aware attention~\cite{liu2018lstm} and trust gates~\cite{liu2018trust}.

GCN-based methods~\cite{Shi_2019_CVPR, Si_2019_CVPR, chen2021multi}, such as ST-GCN~\cite{yan2018spatial}, have proven powerful in modeling the spatio-temporal structure of human motion. Enhancements like Shift-GCN~\cite{Cheng_2020_CVPR} and 2s-AGCN~\cite{Shi_2019_CVPR} further refine these capabilities, making them central to many modern approaches. Our work builds on the ST-GCN~\cite{yan2018spatial} architecture, leveraging its strengths in capturing spatio-temporal dynamics. However, unlike traditional methods that rely on a single skeleton convention, we adapt ST-GCN~\cite{yan2018spatial} to handle multiple skeleton conventions for human action recognition. This allows our model to generalize better across diverse datasets and skeletal representations.

In addition, the reliance on labeled datasets limits the generalization of traditional methods to out-of-domain data. We introduce a multi-skeleton self-supervised learning approach, enabling effective training without extensive labeled data. This enhances robustness, scalability, and applicability to diverse and unseen datasets.

\subsection{Contrastive Learning for Action Recognition}
Contrastive learning is a self-supervised technique that distinguishes between similar and dissimilar pairs of data points to learn robust feature representations. Prominent methods like SimCLR~\cite{chen2020simple} and Momentum Contrast (MoCo)~\cite{He_2020_CVPR} have shown success by leveraging data augmentations and memory modules, particularly when labeled data is scarce.

Incorporating contrastive learning into skeleton-based action recognition presents unique challenges that require tailored strategies. Techniques like AimCLR~\cite{guo2022contrastive} have advanced the field by introducing extreme augmentations specifically designed for 3D skeleton data, significantly enhancing model robustness. We build upon AimCLR~\cite{guo2022contrastive}’s augmentation pipeline to strengthen the backbone of our approach, ensuring that our model effectively handles the complexity of diverse skeleton formats. Similarly, ActCLR~\cite{lin2023actionlet} distinguishes between motion and static regions through adaptive transformations and semantic-aware pooling, pushing the boundaries of action modeling. Our method leverages these advancements, incorporating both AimCLR~\cite{guo2022contrastive}’s rigorous augmentation strategies and ActCLR~\cite{lin2023actionlet}’s nuanced motion analysis, while extending the ST-GCN architecture~\cite{yan2018spatial} to accommodate multiple skeleton conventions. This enables our model to generalize more effectively across diverse datasets and skeletal representations.

While AimCLR~\cite{guo2022contrastive} and ActCLR~\cite{lin2023actionlet} focus on enhancing robustness through extreme augmentations and adaptive transformations, other techniques like CrosSCLR~\cite{Li_2021_CVPR} have ensured consistency across modalities using a MoCov2~\cite{He_2020_CVPR}-based ST-GCN~\cite{yan2018spatial} network. However, CrosSCLR~\cite{Li_2021_CVPR} remains limited to a single skeleton convention. In contrast, our method’s integration of multiple skeleton conventions within a contrastive learning framework represents a significant step forward, enhancing both the robustness and scalability of action recognition models.

By integrating and advancing these state-of-the-art techniques, our approach provides a comprehensive and adaptable solution to the challenges posed by diverse skeletal representations in action recognition. This integration not only addresses the limitations of existing methods but also sets a new standard for robustness and generalization in the field.

%% file: sec/3_method.tex
\section{Method}
\label{sec:method}

Most existing contrastive learning methods for skeleton-based action recognition assume a fixed skeleton format~\cite{Li_2021_CVPR, guo2022contrastive, lin2023actionlet}. This assumption limits their ability to generalize across datasets with different joint definitions, body-part coverage, or connectivity structures. As a result, such methods perform suboptimally when applied to real-world data exhibiting diverse skeletal conventions.

Our Multi-Skeleton Contrastive Learning (MS-CLR), as shown in Fig.~\ref{fig:method_overview}, integrates multiple skeleton conventions into the contrastive learning pipeline, effectively leveraging the varying degrees of information encoded in different parametrizations.

While our training approach can be incorporated into any contrastive learning framework, we implement MS-CLR using the Spatiotemporal Graph Convolutional Network (ST-GCN)~\cite{yan2018spatial} together with MoCov2~\cite{He_2020_CVPR}-based methods, including CrossCLR~\cite{Li_2021_CVPR} and AimCLR~\cite{guo2022contrastive}.

We first detail how we model spatiotemporal evolution of human joints using ST-GCN in Sec.~\ref{sec:st-gcn}. Sec.~\ref{sec:unified-skeleton} describes the integration of multiple skeleton formats into our unified skeleton pose parametrization. Sec.~\ref{sec:contrastive-loss} introduces the contrastive training loss, and Sec.~\ref{sec:integration} outlines how MS-CLR is integrated into existing pipelines.

\subsection{Spatio-Temporal Human Joints via ST-GCN}
\label{sec:st-gcn}

The Spatio-Temporal Graph Convolutional Network (ST-GCN)~\cite{yan2018spatial} serves as the foundation of our approach. The initial step involves constructing a graph $G = \{V, E\}$, where $V$ represents the set of joints across all frames, and $E$ captures the connections between these joints, both within a single frame and across consecutive frames.

In the spatial dimension, edges are defined by joint locations, which represent the anatomical structure of the human body according to the chosen skeleton convention. For instance, joints may include key points such as shoulders, elbows, and knees, with edges corresponding to the connections between these points. Temporally, edges connect the same joint across different frames, effectively capturing the motion of each joint over time. The edge set $E$ is thus composed of two subsets: $E_S = \{v_{ti}v_{tj} \mid (i, j) \in H\}$, where $H$ defines the intra-skeleton edges specific to the chosen skeleton format, and $E_F = \{v_{ti}v_{(t+1)i}\}$, which connects joints across time.

From this graph structure, we construct the adjacency matrix $\mathbf{A} = \left( a_{ij} \right)$, a binary matrix where each element $a_{ij}$ is defined as
\begin{equation}
    a_{ij} = 
    \begin{cases}
        1 & \text{if an edge between $v_i$ and $v_j$ exists} \\
        0 & \text{otherwise}.
    \end{cases}
\end{equation}

The neighboring joints of a given joint $v_{ti}$ are defined as:
\begin{equation}
    B(v_{ti}) = \{ v_{qj} \mid d(v_{tj}, v_{ti}) \leq K, \ |q - t| \leq \lfloor \Gamma / 2 \rfloor \},
\end{equation}
where $K$ and $\Gamma$ represent the spatial and temporal kernel sizes, and $d$ measures the discrete vertex-to-vertex hops along graph edges. These neighbors are further categorized into three subsets: the root node, centripetal group (closer to the skeleton's center of gravity), and centrifugal group (further from the center).

\input{figures/skeletons/skeletons}

Spatiotemporal graph convolutions~\cite{bronstein2017geometric} are then applied following the ideas of ST-GCN~\cite{yan2018spatial} with
\begin{equation}
f_{\text{out}}(v_{ti}) = \sum_{v_{tj} \in B(v_{ti})} \frac{1}{Z_{ti}(v_{tj})} f_{\text{in}}(v_{tj}) \cdot w(l_{ti}(v_{tj})),
\end{equation}
where:
\begin{itemize}
    \item $f_{\text{in}}(v_{tj}) \in \mathbb{R}^{n}$ is the input feature at joint $v_{tj}$,
    \item $f_{\text{out}}(v_{ti}) \in \mathbb{R}^{n}$ is the output feature at joint $v_{ti}$,
    \item $Z_{ti}(v_{tj})$ is a normalizing term based on neighborhood size,
    \item $w(l_{ti}(v_{tj}))$ is a learnable weight determined by the structural label $l_{ti}(v_{tj})$.
\end{itemize}

As a result, $f_{\text{out}}(v_{ti})$ encodes both spatial and temporal dynamics and is passed to downstream layers for action recognition.

\subsection{Unified Skeleton ST-GCN Adaptation}
\label{sec:unified-skeleton}

MS-CLR leverages diverse skeleton conventions that vary in joint count, anatomical detail, and structural design, as shown in Fig.~\ref{fig:skeleton_comparison}. SMPL~\cite{loper2023smpl} provides a compact full-body model with 24 joints, widely adopted in graphics and animation. SMPL-X~\cite{pavlakos2019expressive}, in our setting, extends SMPL by including hand and facial articulation, yielding a 42-joint representation. Berkeley MHAD~\cite{ofli2013berkeley} defines a 43-joint layout tailored for motion capture, with extensive coverage of upper and lower limbs, hands, and feet. Kinectv2~\cite{zhang2012microsoft} offers a streamlined 25-joint format optimized for real-time tracking, with less emphasis on fine-grained detail. This diversity provides useful structural variation but presents challenges for learning consistent representations.

To address this, we introduce a unified pose representation. While existing methods like CrossCLR~\cite{Li_2021_CVPR} operate on a single native format, our approach extracts multiple auxiliary skeleton formats from RGB videos using MeTRAbs~\cite{Sarandi_2023_WACV}, a pose estimator trained on 28 datasets that can produce 23 distinct skeleton conventions from the same input image.

Skeletons differ in joint count, connectivity, and topology. To support consistent batching, we zero-pad pose tensors and their adjacency matrices to match the largest skeleton size:
\begin{equation}
    V_{\text{max}} = \max(V_s), \quad s \in \{ 1, \ldots, S \},
\end{equation}
where $S$ is the number of skeleton conventions. Each skeleton sequence $p_i^s$ is stored in $\mathbb{R}^{C \times V_{\text{max}} \times T \times P}$, where $C$ is channel dimension, $T$ is the number of frames, and $P$ is the number of people.

Padded joints and edges are masked out and do not contribute to training. This allows unified batch processing across skeletons while preserving each skeleton’s original topology and graph connectivity.

We extend the ST-GCN adjacency matrix $\mathbf{A}$ to a set $\bar{\mathbf{A}} = \left( \mathbf{A}_1, \ldots, \mathbf{A}_S \right)$, where each $\mathbf{A}_s$ corresponds to skeleton format $s$ and is constructed using joint connections $H_s$.

The generalized multi-skeleton graph convolution becomes:
\begin{equation}
f_{\text{out}}(v_{ti}^{s}) = \sum_{v_{tj} \in B_s(v_{ti})} \frac{1}{Z_{ti}(v_{tj})} f_{\text{in}}(v_{tj}) \cdot w(l_{ti}(v_{tj})),
\end{equation}
preserving the ST-GCN structure while enabling format-aware learning across skeleton types.

While we demonstrate MS-CLR using four skeleton formats in our experiments, the framework is not limited to these configurations. It is compatible with any pose estimation method that produces joint-based sequences and supports arbitrary combinations of skeleton conventions. The unified input representation and format-specific adjacency structure allow seamless integration of new formats without requiring changes to the model architecture.

\subsection{Training via Multi-Skeleton Contrastive Loss}
\label{sec:contrastive-loss}

MS-CLR generates embeddings from each skeleton format, which are contrasted using a temperature-scaled InfoNCE loss~\cite{oord2018representation}:
\begin{equation}
    \mathcal{L} = -\log \frac{\exp(q \cdot k / \tau)}{\exp(q \cdot k / \tau) + \sum_{i=1}^{N} \exp(q \cdot m_i / \tau)},
    \label{eq:loss}
\end{equation}
where $q$ is the query, $k$ is the positive key from another skeleton format of the same sample, and $\{m_i\}$ are negatives in the memory bank.

We follow the MoCo~\cite{He_2020_CVPR} design: query and key encoders share the same architecture but are maintained separately. The key encoder is updated via an exponential moving average of the query encoder's weights, ensuring smoother updates and stabilizing training. A shared memory bank stores embeddings across all skeleton formats.

This loss encourages consistency between views of the same action under different skeleton conventions while enforcing discrimination from unrelated samples. Unlike standard augmentations that simulate low-level perturbations, variation across skeleton formats reflects different structural assumptions in pose estimation. SMPL-X~\cite{pavlakos2019expressive} includes fine-grained articulation in the hands and face, Kinectv2~\cite{zhang2012microsoft} provides a compact representation with coarser body coverage, and Berkeley MHAD~\cite{ofli2013berkeley} emphasizes limb-level detail. Aligning representations across such structurally diverse formats encourages the model to focus on the core motion dynamics while discarding format-specific biases. As a result, the model learns representations that are both structure-invariant and more transferable across datasets and robust to variation in pose estimation systems.

\subsection{Integration of MS-CLR into Existing Pipelines}
\label{sec:integration}

We integrate MS-CLR into the AimCLR~\cite{guo2022contrastive} pipeline as our base framework for self-supervised learning. AimCLR applies strong spatial and temporal augmentations, including Shear, Crop, Flip, Gaussian Noise, and Blur, to encourage robust representation learning from skeleton sequences.

To extend AimCLR into a multi-skeleton setting, we apply the same set of augmentations independently to each skeleton format and enforce alignment between their representations using the contrastive loss in Eq.~\eqref{eq:loss}. This enables the model to learn structure-invariant representations from diverse skeleton layouts.

Variation in skeleton format serves as a structural augmentation signal. Training with formats that differ in joint count, body coverage, and anatomical detail improves robustness to pose estimator design and enhances generalization across datasets.

\subsection{Skeleton Specific Classifier Ensembling}
\label{sec:skeleton-ensembling}

In addition to learning shared representations through contrastive training, MS-CLR supports a skeleton specific classifier ensembling strategy. Since the model processes multiple skeleton conventions during pretraining, we preserve format-specific features by maintaining separate linear classifiers for each skeleton type during evaluation.

Each classifier is trained independently on the frozen backbone representations corresponding to a single skeleton convention. At test time, the final prediction is computed by averaging the softmax scores produced by each classifier. This ensemble approach exploits the complementary nature of different skeleton formats: coarser formats such as Kinectv2~\cite{zhang2012microsoft} emphasize global posture, while higher-resolution formats like SMPL-X~\cite{pavlakos2019expressive} capture fine-grained details in hands and face.

This strategy improves recognition accuracy without requiring any architectural changes or retraining, and it can be applied to any number of skeleton formats used during pretraining.

%% file: figures/skeletons/skeletons.tex
\begin{figure}[t]
  \centering
  \captionsetup[subfigure]{justification=centering}
  \begin{subfigure}[t]{0.3\linewidth}
    \centering
    \includegraphics[width=\linewidth]{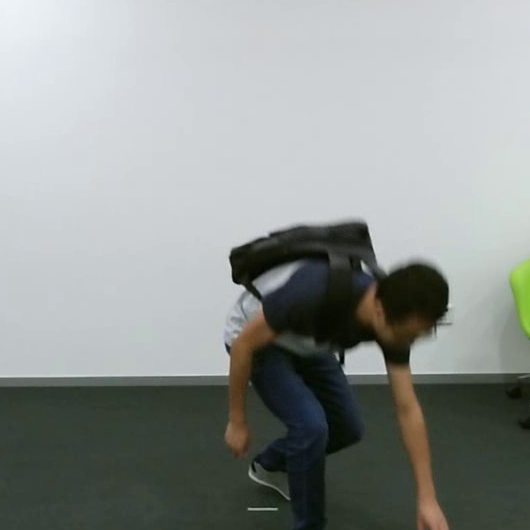}
    \caption{RGB Input}
    \label{fig:skel_rgb}
  \end{subfigure}
  \begin{subfigure}[t]{0.3\linewidth}
    \centering
    \includegraphics[width=0.8\linewidth]{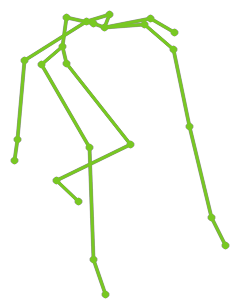}
    \caption{SMPL}
    \label{fig:skel_smpl}
  \end{subfigure}
  \begin{subfigure}[t]{0.3\linewidth}
    \centering
    \includegraphics[width=0.8\linewidth]{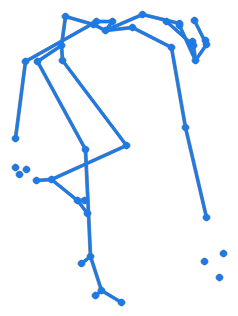}
    \caption{SMPL-X}
    \label{fig:skel_smplx}
  \end{subfigure}

  \vspace{1em}

  \begin{subfigure}[t]{0.3\linewidth}
    \centering
    \includegraphics[width=0.8\linewidth]{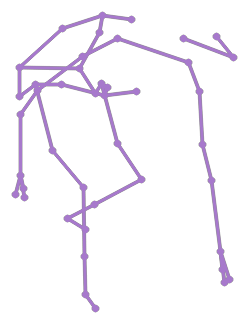}
    \caption{Berkeley MHAD}
    \label{fig:skel_berkeley}
  \end{subfigure}
  \begin{subfigure}[t]{0.3\linewidth}
    \centering
    \includegraphics[width=0.8\linewidth]{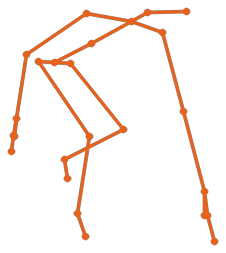}
    \caption{Kinectv2}
    \label{fig:skel_kinect}
  \end{subfigure}

  \caption{
Visualization of example skeleton formats used in MS-CLR. From a single RGB frame (a), we can extract four distinct skeleton representations: (b) SMPL~\cite{loper2023smpl} (24 joints), (c) SMPL-X~\cite{pavlakos2019expressive} (42 joints, including hands and face), (d) Berkeley MHAD~\cite{ofli2013berkeley} (43 joints), and (e) Kinectv2~\cite{zhang2012microsoft} (25 joints). These formats differ in joint density, anatomical coverage, and structural layout.
}
  \label{fig:skeleton_comparison}
\end{figure}

%% file: sec/4_experiments.tex
\section{Experiments}
\label{sec:experiments}

We conduct extensive experiments on two widely used benchmarks for skeleton-based action recognition: NTU RGB+D 60~\cite{shahroudy2016ntu, liu2019ntu} and NTU RGB+D 120~\cite{liu2019ntu}. Our evaluation focuses on assessing the effectiveness of the proposed multi-skeleton contrastive learning (MS-CLR) framework under various configurations and training regimes.

\begin{table}[t]
  \centering
  \caption{Comparison of action recognition results between AimCLR and our multi-skeleton variant, MS-AimCLR, on the NTU RGB+D dataset. Streams: `j` - joint, `m` - motion, `b` - bone. \customddag~denotes our multi-skeleton ensemble model.}
\resizebox{\columnwidth}{!}{
\begin{tabular}{llcccc}
    \toprule
    Models & Stream & \multicolumn{2}{c}{NTU 60 (\%)} & \multicolumn{2}{c}{NTU 120 (\%)} \\
    \cmidrule(lr){3-4} \cmidrule(lr){5-6}
    & & xview & xsub & xset & xsub \\
    \midrule
    AimCLR~\cite{guo2022contrastive} & j & 79.7 & 74.3 & 63.4 & 63.4 \\
    MS-AimCLR & j & \textbf{83.0} & \textbf{76.1} & \textbf{67.6} & \textbf{63.9} \\
    \midrule
    AimCLR~\cite{guo2022contrastive} & m & 70.6 & 66.8 & 54.4 & 57.3 \\
    MS-AimCLR & m & \textbf{80.4} & \textbf{73.1} & \textbf{62.6} & \textbf{60.7} \\
    \midrule
    AimCLR~\cite{guo2022contrastive} & b & 77.0 & 73.2 & 63.4 & 62.9 \\
    MS-AimCLR & b & \textbf{82.2} & \textbf{76.1} & \textbf{67.3} & \textbf{64.0} \\
    \midrule
    3s-AimCLR~\cite{guo2022contrastive} & j+m+b & 83.8 & 78.9 & 68.8 & 68.2 \\
    3s-MS-AimCLR & j+m+b & 86.7 & 80.9 & 70.1 & 68.3 \\
    3s-MS-AimCLR~\customddag & j+m+b & \textbf{94.2} & \textbf{88.0} & \textbf{75.1} & \textbf{75.3} \\
    \bottomrule
  \end{tabular}
  }  
  \label{tab:ntu_comparison_aimclr}
\end{table}

\subsection{Datasets and Training Setup}

\textbf{NTU RGB+D 60 (NTU 60)}~\cite{shahroudy2016ntu} contains 56,880 action clips covering 60 categories, performed by 40 subjects and captured from three fixed horizontal viewpoints. The skeletal annotations follow the Kinectv2~\cite{zhang2012microsoft} convention, comprising 25 joints. We report results under two standard evaluation protocols: \textit{cross-subject (X-Sub)}, where 20 subjects are used for training and the remaining 20 for testing, and \textit{cross-view (X-View)}, which splits data based on camera perspectives, with cameras 2 and 3 used for training and camera 1 for testing.

\textbf{NTU RGB+D 120 (NTU 120)}~\cite{liu2019ntu} is an extended version of NTU 60, introducing 60 additional action classes and increasing the total number of clips to 114,480. These sequences are performed by 106 subjects under 32 different setup conditions. Two protocols are used: \textit{cross-subject (X-Sub)}, which divides the 106 subjects into disjoint training and test groups, and \textit{cross-setup (X-Set)}, which splits the data based on recording setups, using even-numbered setups for training and odd-numbered ones for testing.

\textbf{Training Configuration.} All skeleton sequences are linearly interpolated to 50 frames. Our encoder backbone is a multi-skeleton variant of ST-GCN, which accepts inputs from multiple skeleton conventions by using a unified joint representation (via zero-padding) and format-specific adjacency matrices. The feature encoder produces 256-dimensional embeddings, which are projected to 128 dimensions through an MLP head for contrastive learning.

\textbf{Pretraining Strategy.} During unsupervised pretraining, skeletons from different conventions are extracted from the same RGB sequence using independent pose estimators. These are treated as alternative views in the contrastive loss. In our default configuration, we use two complementary skeleton formats: Kinectv2 (moderate-resolution, native to NTU) and SMPL-X~\cite{pavlakos2019expressive} (high-resolution, full-body). This combination enables the model to learn features that are robust to structural variation in topology, resolution, and body-part coverage.

We adopt stochastic gradient descent with momentum 0.9 and weight decay of 1e-4. Models are trained for 300 epochs, with an initial learning rate of 0.1 reduced to 0.01 at epoch 250. Since each clip may yield multiple skeleton formats, the number of iterations per epoch is scaled proportionally to preserve the effective sampling rate per format.

\textbf{Multi-Stream Inputs.} Following prior work~\cite{guo2022contrastive, lin2023actionlet}, we construct three input streams: joint, bone, and motion. For multi-stream models, we apply late fusion with fixed weights of [0.6, 0.6, 0.4] respectively. All training and evaluation are conducted in PyTorch 2.2.1 on a single NVIDIA RTX 4090 GPU.

\begin{table}[t]
  \centering
    \caption{Comparison of action recognition results on the NTU RGB+D 60 dataset across various unsupervised learning approaches. The table contrasts both single-stream and three-stream methods. \customddag~indicates our multi-skeleton ensemble models.}
  \resizebox{\columnwidth}{!}{
  \begin{tabular}{lccc}
    \toprule
    Models & Architecture & xview & xsub \\
    \midrule
    \textbf{Single-stream:} \\
    SeBiReNet~\cite{nie2020unsupervised} & SeBiReNet & 79.7 & - \\
    ISC~\cite{thoker2021skeleton} & GCN \& GRU & 78.6 & 76.3 \\
    AimCLR~\cite{guo2022contrastive} & GCN & 79.7 & 74.3 \\
    CMD~\cite{mao2022cmd} & GRU & 81.3 & 76.8 \\
    GL-Transformer~\cite{kim2022global} & Transformer & 83.8 & 76.3 \\
    CPM~\cite{zhang2022contrastive} & GCN & 84.9 & 78.7 \\
    ActCLR~\cite{lin2023actionlet} & GCN & 86.7 & 80.9 \\
    MS-AimCLR~\customddag \ (Ours) & GCN & \textbf{93.2} & \textbf{86.6} \\
    \midrule
    \textbf{Three-stream:} \\
    3s-Colorization~\cite{yang2021skeleton} & DGCNN & 83.1 & 75.2 \\
    3s-CrosSCLR~\cite{Li_2021_CVPR} & GCN & 83.4 & 77.8 \\
    3s-AimCLR~\cite{guo2022contrastive} & GCN & 83.8 & 78.9 \\
    3s-CMD~\cite{mao2022cmd} & GRU & 85.0 & 79.9 \\
    3s-SkeleMixCLR~\cite{chen2022contrastive} & GCN & 87.1 & 82.7 \\
    3s-CPM~\cite{zhang2022contrastive} & GCN & 87.0 & 83.2 \\
    3s-ActCLR~\cite{lin2023actionlet} & GCN & 88.8 & 84.3 \\
    3s-MS-AimCLR~\customddag \ (Ours) & GCN & \textbf{94.2} & \textbf{88.0} \\
    \bottomrule
  \end{tabular}
  }
  \label{tab:ntu60_comparison}
\end{table}

\begin{table}[t]
  \centering
    \caption{Comparison of action recognition results on the NTU RGB+D 120 dataset across various unsupervised learning approaches. The table contrasts both single-stream and three-stream methods. \customddag~indicates our multi-skeleton ensemble models.}
  \resizebox{\columnwidth}{!}{
  \begin{tabular}{lccc}
    \toprule
    Models & Architecture & xset & xsub \\
    \midrule
    \textbf{Single-stream:} \\
    AS-CAL~\cite{rao2021augmented} & LSTM & 49.2 & 48.6 \\
    AimCLR~\cite{guo2022contrastive} & GCN & 63.4 & 63.4 \\
    CMD~\cite{mao2022cmd} & GRU & 66.0 & 65.4 \\
    GL-Transformer~\cite{kim2022global} & Transformer & 68.7 & 66.0 \\
    CPM~\cite{zhang2022contrastive} & GCN & 69.6 & 68.7 \\
    ActCLR~\cite{lin2023actionlet} & GCN & 70.5 & 69.0 \\
    MS-AimCLR~\customddag \ (Ours) & GCN & \textbf{73.8} & \textbf{74.3} \\
    \midrule
    \textbf{Three-stream:} \\
    3s-CrosSCLR~\cite{Li_2021_CVPR} & GCN & 66.7 & 67.9 \\
    3s-AimCLR~\cite{guo2022contrastive} & GCN & 68.8 & 68.2 \\
    3s-CMD~\cite{mao2022cmd} & GRU & 69.6 & 69.1 \\
    3s-SkeleMixCLR~\cite{chen2022contrastive} & GCN & 70.7 & 70.5 \\
    3s-CPM~\cite{zhang2022contrastive} & GCN & 74.0 & 73.0 \\
    3s-ActCLR~\cite{lin2023actionlet} & GCN & \textbf{75.7} & 74.3 \\
    3s-MS-AimCLR~\customddag \ (Ours) & GCN & 75.1 & \textbf{75.3} \\
    \bottomrule
  \end{tabular}
  }
  \label{tab:ntu120_comparison}
\end{table}

\subsection{Evaluation and Comparison}

\textbf{Linear Evaluation Protocol.} 
Following common self-supervised evaluation protocols~\cite{guo2022contrastive, lin2023actionlet}, we assess representation quality using a linear evaluation setup. After unsupervised pretraining, we freeze the encoder and train a linear classifier for 100 epochs on labeled data with batch size 128. The learning rate is initialized at 3.0 and decayed by a factor of 0.1 at epoch 80. For consistency with prior benchmarks, all models are evaluated on the NTU RGB+D datasets using the Kinectv2 skeleton convention as the reference format at test time.

We report results in both single-stream and multi-stream settings:
\begin{itemize}
    \item \textbf{Single-stream:} Models are evaluated using only one modality (joint, motion, or bone).
    \item \textbf{Multi-stream (3s):} All three modalities are fused via weighted averaging for final prediction.
\end{itemize}

The results in Table~\ref{tab:ntu_comparison_aimclr} show that MS-AimCLR consistently improves upon the baseline AimCLR across all streams and datasets. Notably, performance gains are most pronounced in the motion stream, highlighting the sensitivity of temporal cues to structural skeleton variation. On NTU 60, our variant improves joint-based accuracy by +3.3 points in X-View and +1.8 in X-Sub, while motion-based accuracy improves by +9.8 and +6.3 points respectively. The improvements are also evident on NTU 120, where MS-AimCLR outperforms AimCLR by +4.2 points (X-Set, joint) and +5.3 points (motion).

In the multi-stream setting, 3s-MS-AimCLR surpasses 3s-AimCLR by +2.9 in X-View and +2.0 in X-Sub, setting new state-of-the-art results. This suggests that skeleton diversity acts as a natural structural augmentation, enhancing the encoder’s ability to generalize across viewpoints and subjects.

\textbf{Multi-Skeleton Ensembling.}
We also explore a skeleton-specific ensemble strategy, where linear classifiers are trained independently on embeddings from each skeleton convention. Final predictions are averaged across classifiers. This approach exploits the complementarity between formats, where some formats emphasize global posture and others provide finer-grained articulation.

As shown in Tables~\ref{tab:ntu_comparison_aimclr}, \ref{tab:ntu60_comparison}, and \ref{tab:ntu120_comparison}, ensembling further boosts performance. On NTU 60, 3s-MS-AimCLR~\customddag achieves 94.2 in X-View and 88.0 in X-Sub, surpassing previous methods by significant margins. Similar trends are observed on NTU 120, with absolute improvements of +3 to +4 points over prior best results. These gains are especially prominent in complex actions involving detailed limb or hand articulation, where higher-resolution formats like SMPL-X provide additional discriminative cues.

\begin{figure}[t]
  \centering
  \includegraphics[width=\columnwidth]{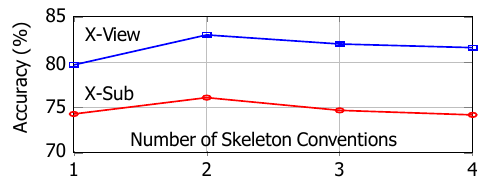}
  \caption{Impact of skeleton diversity on NTU-60 accuracy. Accuracy improves most when combining two complementary formats: Kinectv2~\cite{zhang2012microsoft} and SMPL-X~\cite{pavlakos2019expressive}. Adding further formats such as Berkeley MHAD~\cite{ofli2013berkeley} and SMPL~\cite{loper2023smpl} yields limited gains, suggesting diminishing returns from structurally similar conventions.}
  \label{skeletonConv}
\end{figure}

\begin{table}[t]
  \centering
  \caption{Component analysis of the MS-CLR framework on NTU-60.}
  \resizebox{\columnwidth}{!}{
  \begin{tabular}{ccccc}
    \toprule
    Augmentations & Multi-Skeleton & Ensemble & \multicolumn{2}{c}{NTU 60 Accuracy (\%)} \\
    \cmidrule(lr){4-5}
    & & & xsub & xview \\
    \midrule
    \checkmark & & & 79.7 & 74.3 \\
    & \checkmark & & 53.2 & 49.1 \\
    \checkmark & \checkmark & & 83.0 & 76.1 \\
    \checkmark & \checkmark & \checkmark & \textbf{93.2} & \textbf{86.6} \\
    \bottomrule
  \end{tabular}
  }
  \label{tab:augmentation_multiskeleton}
\end{table}

\subsection{Ablation Study}

\textbf{Effect of the Number of Skeleton Formats.} 
We evaluate the contribution of skeleton diversity during contrastive pretraining by varying the number and type of formats used. As shown in Figure~\ref{skeletonConv}, model accuracy increases with the inclusion of additional skeleton conventions, with the most significant gains observed when two structurally distinct formats are combined. In particular, pairing Kinectv2~\cite{zhang2012microsoft} and SMPL-X~\cite{pavlakos2019expressive} yields the strongest improvements, reflecting their complementary anatomical coverage. Further addition of Berkeley MHAD~\cite{ofli2013berkeley} and SMPL~\cite{loper2023smpl} results in marginal gains, suggesting diminishing returns when format redundancy increases.

This analysis highlights that structural variety is key to learning generalizable representations. Kinectv2 captures coarse posture and global body dynamics, while SMPL-X introduces fine-grained joint detail, particularly in hands and face. Together, they provide complementary motion cues across resolutions and topologies. Conversely, closely related formats (e.g., SMPL and SMPL-X) introduce less novel structural variation, limiting their added value. This underscores the importance of choosing diverse and complementary skeletons to guide the model toward structure-aware representations rather than relying solely on redundancy.

\begin{figure}[t]
  \centering
  \includegraphics[width=\columnwidth]{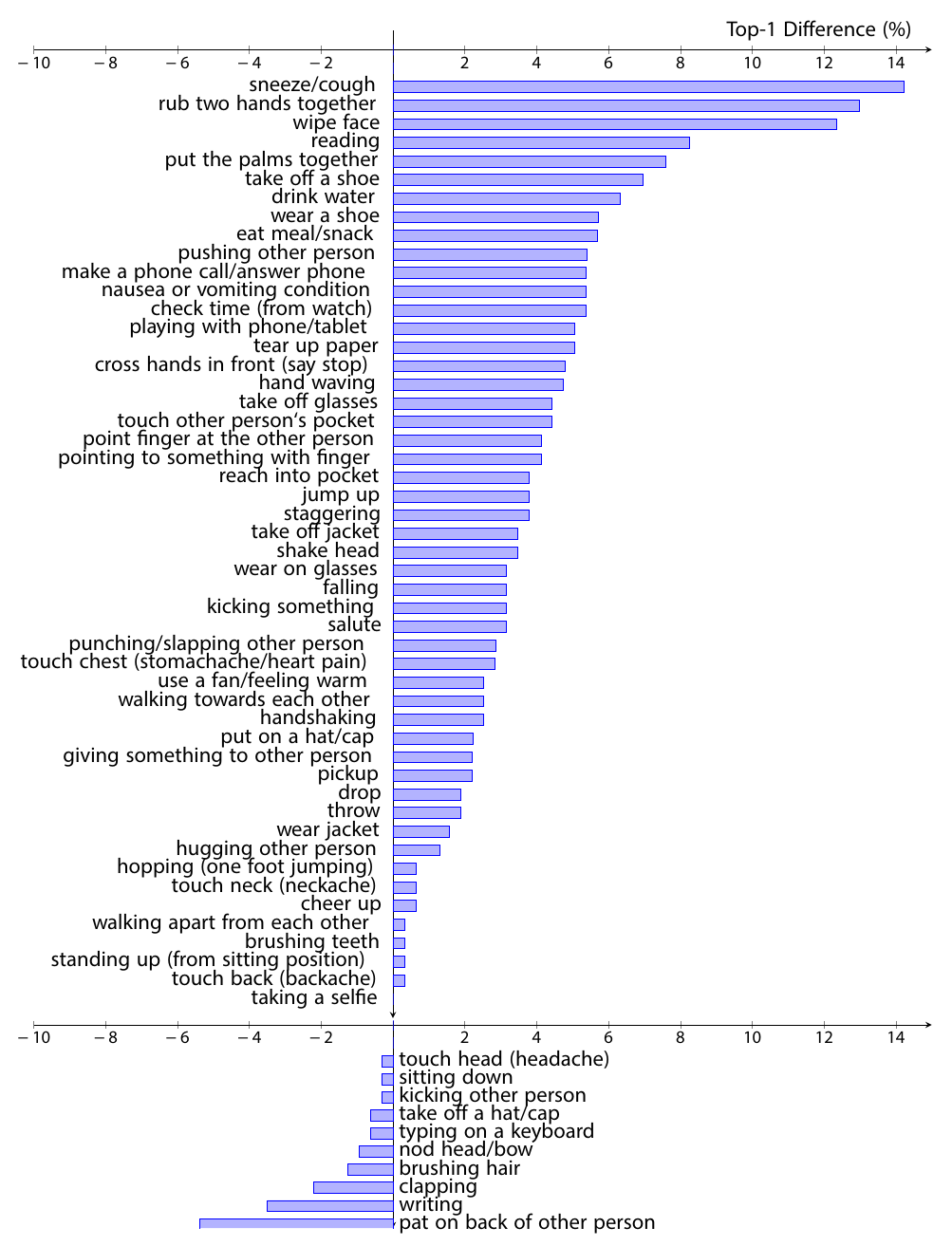}
  \caption{ Class-wise Top-1 accuracy improvement on NTU-60 when using MS-CLR (3s-MS-CrosSCLR) over its single-skeleton counterpart (3s-CrosSCLR~\cite{Li_2021_CVPR}). Positive bars indicate action classes that benefit from skeleton diversity, with larger gains in fine-grained actions such as "wear a shoe" and "answer phone."}
  \label{sortedDiff}
\end{figure}

\textbf{Component-Wise Analysis.}
To assess the contribution of key components in MS-CLR, we perform a systematic ablation study isolating augmentations, skeleton diversity, and ensembling. Results are presented in Table~\ref{tab:augmentation_multiskeleton}.

Applying only standard skeleton augmentations (temporal jitter, crop, shear, Gaussian blur) yields solid performance, consistent with prior self-supervised methods. In contrast, using multi-skeleton contrastive learning without augmentations leads to poor performance. This suggests that format diversity alone cannot substitute for low-level perturbations. However, when both components are combined, performance improves by +3.3 (X-Sub) and +1.8 (X-View) over the augmentation-only baseline, confirming that structural augmentation and conventional perturbations offer complementary benefits.

Ensembling further amplifies performance, increasing accuracy to 93.2 (X-Sub) and 86.6 (X-View). This reflects the effectiveness of aggregating skeleton-specific representations, where different formats capture distinct action semantics. The result supports the design of MS-CLR as a modular framework where contrastive invariance, architectural flexibility, and ensemble diversity contribute synergistically.

\textbf{Per-Class Performance Analysis.}
To better understand which action categories benefit most from multi-skeleton learning, we compare the per-class Top-1 accuracy between MS-CLR and a single-format baseline (3s-CrosSCLR~\cite{Li_2021_CVPR}) on NTU 60. As shown in Figure~\ref{sortedDiff}, MS-CLR provides notable gains in actions that require fine-grained articulation or subtle limb movement, such as ``wear a shoe,'' ``answer phone,'' and ``tear up paper.'' These categories involve nuanced hand or finger motions, which are better captured by high-resolution skeletons like SMPL-X~\cite{pavlakos2019expressive} that include detailed joints in the hands and face.

We also observe improvements in full-body motion categories like ``walking apart'' and ``falling,'' which require consistent modeling of global posture and coordination. These gains highlight the complementary strengths of different skeleton conventions: SMPL-X offers detailed articulation, Kinectv2~\cite{zhang2012microsoft} provides a stable coarse body layout, and Berkeley MHAD~\cite{ofli2013berkeley} emphasizes limb structure. Contrasting across such formats helps the model capture task-relevant motion while being invariant to estimator-specific biases, leading to more transferable and structure-aware features.

%% file: sec/5_conclusion.tex
\section{Method Limitations and Data Considerations}
\label{sec:limitations}
Despite significant overall improvements in action recognition and the ability of MS-CLR to transfer skeleton-sensitive action knowledge such as foot-focused actions across datasets, we also observed some slight drop in performance on a few action classes (cf. Fig.~\ref{sortedDiff}). These involve actions that require accurate finger-joint estimation, where a specialized skeleton model trained solely on fine-grained data may perform better. While MS-CLR focuses on motion patterns rather than detailed visual data analysis, it inherently reduces identity exposure and addresses some data privacy concerns. However, action recognition remains a sensitive topic where data storage, anonymization, and GDPR compliance are critical, and our method shares the same potential for unauthorized surveillance as other approaches in the field.

\section{Conclusion}
\label{sec:conclusion}
We present MS-CLR, a contrastive learning framework that leverages structural diversity across multiple skeleton conventions to learn more robust and transferable action representations. By unifying anatomically diverse skeleton formats within a shared graph-based architecture, our method captures complementary motion cues that single-format models often miss. Experiments on NTU RGB+D 60 and 120 demonstrate that MS-CLR consistently improves performance across tasks and modalities, with further gains achieved through skeleton-specific ensembling. Beyond improved accuracy, MS-CLR offers a scalable way to integrate pose information from structurally distinct sources, enabling broader applicability in real-world action recognition. We hope our approach and accompanying code release will support future research in structure-aware representation learning.